\def\year{2021}\relax
\documentclass[letterpaper]{article} 
\usepackage{aaai21}  
\usepackage{times}  
\usepackage{helvet} 
\usepackage{courier}  
\usepackage[hyphens]{url}  
\usepackage{graphicx} 
\usepackage{natbib}  
\usepackage{caption} 
\usepackage{hyperref}
\usepackage{amsfonts}
\usepackage{multirow}
\usepackage{booktabs, microtype}
\usepackage{multicol}

\nocopyright
\title{\textbf{EC-GAN: Low-Sample Classification using Semi-Supervised 
\\ Algorithms and GANs}}
\author {
    Ayaan Haque \\
}

\affiliations{
    Saratoga High School, Saratoga, CA, USA \\
    ayaanzhaque@gmail.com
}

\begin{document}

\maketitle

\begin{abstract}

Semi-supervised learning has been gaining attention as it allows for performing image analysis tasks such as classification with limited labeled data. Some popular algorithms using Generative Adversarial Networks (GANs) for semi-supervised classification share a single architecture for classification and discrimination. However, this may require a model to converge to a separate data distribution for each task, which may reduce overall performance. While progress in semi-supervised learning has been made, less addressed are small-scale, fully-supervised tasks where even unlabeled data is unavailable and unattainable. We therefore, propose a novel GAN model namely \emph{External Classifier GAN (EC-GAN)}, that utilizes GANs and semi-supervised algorithms to improve classification in fully-supervised regimes. Our method leverages a GAN to generate artificial data used to supplement supervised classification. More specifically, we attach an external classifier, hence the name EC-GAN, to the GAN’s generator, as opposed to sharing an architecture with the discriminator. Our experiments demonstrate that EC-GAN's performance is comparable to the shared architecture method, far superior to the standard data augmentation and regularization-based approach, and effective on a small, realistic dataset. \footnote{Code is available at \href{https://github.com/ayaanzhaque/EC-GAN}{https://github.com/ayaanzhaque/EC-GAN}}

\end{abstract}

\section{Introduction}

Deep learning has grown in interest in recent years, but its growth and effectiveness has relied on the availability of large, labeled datasets. Semi-supervised learning combines a small amount of labeled data with a large amount of unlabeled data during training \citep{chapelle2009semi}. As a result, semi-supervised learning has received much attention as an alternative to fully-labeled datasets, and many works and methods have been proposed \cite{chen2020naivestudent, sohn2020fixmatch, haque2020multimix, laine2017temporal, imran2020partly, sun2019unsupervised, imran2019semi}. In many real-world tasks, there are large datasets with only a small subset of labeled data, as annotations require domain expertise, are expensive, and are time consuming. Specifically for deep learning models and neural networks, large datasets are often required to train an accurate network. Semi-supervised learning aims to address this concern and aid the growth of deep learning by maximizing knowledge gains from datasets. With the goal in mind of maximizing datasets, this paper addresses a more severe learning scenario: fully-supervised classification tasks with restricted and low-sample datasets.

While many datasets have an abundance of unlabeled data, there are conversely a significant amount of datasets where even unlabeled data is unavailable or unattainable. For these tasks, collecting data is more challenging than labeling data. For example, in the medical domain, in order to acquire a chest x-ray, a radiologist or similar specialist must properly administer an x-ray. Moreover, this requires a patient who is willing to take an x-ray. Additionally, in order for the image to be valuable for a certain classification task, it must be representative of the disease of interest. However, labeling the samples may not necessarily be as difficult, as if a chest x-ray is taken, it is subsequently labeled and annotated, as x-rays are taken for diagnostic purposes to begin with. These issues result in low-sample datasets, restricting the effectiveness of deep-learning for these tasks. These datasets can greatly benefit from any kind of new data, even unlabeled or artificial. Our method aims to improve classification in these specific tasks.

Deep generative models can generate artificial samples by learning an underlying data distribution. Recently, Generative Adversarial Networks \cite{gan} have emerged as one of the most effective generative models. Other popular generative models include Variational Autoencoder \cite{kingma2013auto}, or VAEs, which provide a probabilistic method for describing an observation in latent space, often by learning a low-dimensional latent representation of the data. VAEs contain two modules, an encoder and a decoder, and this encoder-decoder framework has become heavily utilized in generative modeling. Many methods have recently been proposed using VAEs for image generation \cite{imran2021multi, vahdat2020nvae, sagar2020generate} VAEs however differ from GANs, which are the generative models used in our method. Most generative models are unsupervised and do not require real annotations, therefore improving the ability to train effective generative algorithms. Our intuition for our method is motivated by the nature of a GAN, which generates realistic images based on a given dataset. We believe there is high value in generating artificial data, specifically for classification. 

We introduce EC-GAN, an algorithm to improve classification using semi-supervised algorithms and Generative Adversarial Networks. GANs have previously been researched and utilized for unsupervised and semi-supervised learning and have seen promising results \cite{dcgan, triplegan, generativemodels, improvedgans, imran2019multi, ccgan, adversarialinference}. However, we apply a GAN to fully-supervised scenarios, specifically restricted data regimes, which to our knowledge has not been researched as heavily. A GAN provides artificial images that are unique from existing images while still resembling real data, and these generated samples can supplement real ones during training.

Our main contributions may be summarized as follows:

\begin{itemize}
\item 
A novel semi-supervised method exploiting generative adversarial networks to aid classification tasks through the use of artificial data.

\item 
An intuitive architectural scheme which separates the networks for classification and discrimination while still maintaining a multi-tasking relationship.

\item 
Comprehensive experimentation specifically on restricted, fully-supervised datasets, both in an academic and real-world setting, paired with an extensive ablation study of each proposed addition.
\end{itemize}

This paper begins by discussing specific related work in the next section. Then we thoroughly detail our approach and algorithm in the subsequent section. Finally, we review our experimentation and discuss the results in the final section.

\section{Related Work}
\label{Related Work}
\textbf{Generative Adversarial Networks}: Generative Adversarial Networks \cite{gan} train two neural networks where a generative model attempts to generate images approximating the distribution of real training samples. Simultaneously, a discriminative network predicts the probability that a generated image is from the real training set. The two models compete with one another, such that the generator eventually produces images resembling real training samples. During training, the generator is updated on predictions of the discriminator in order to produce better images, and the discriminator improves at discriminating images as real or fake. The minimax objective of a GAN can be written as
\begin{equation}
    \mathbb{E}_x[log(D(x))] + \mathbb{E}_z[log(1-D(G(z)))],
\end{equation}
where $\mathbb{E}_x$ is the expected value over all data instances, D($x$) is the discriminator’s predicted probability of an image being real, and $\mathbb{E}_z$ is the expected value of the random inputs into the generator. The minimum of this loss is where $\mathbb{E}_x = \mathbb{E}_z$, which indicates the networks have achieved equilibrium, at which point the generator is creating almost perfect images and the discriminator is left with a 50\% chance of discriminating correctly. We incorporate a GAN in our algorithm to generate images that can supplement fully-supervised classification. 

\textbf{Deep Convolutional GAN}: We use the Deep Convolutional Generative Adversarial Network \cite{dcgan} architecture for the generator and discriminator in our method. DCGAN is a direct extension of the original GAN framework utilizing a convolutional architecture. DCGAN we developed to generate higher quality images, as at the time of its introduction, GAN performance on large and pixel-dense images was comparatively poor. We utilize the DCGAN architecture for image classification in fully-supervised datasets, as opposed to other works that have used the DCGAN for semi-supervised or unsupervised learning.

\textbf{Pseudo-Labeling}: Pseudo-labeling \cite{pseudolabel} aims to train on unlabeled data by leveraging labeled data. Pseudo-labeling is a form of self-training, where a model trained in a supervised fashion will then create artificial labels for unlabeled images in order to learn from the unlabeled images. To achieve this, the model simply chooses the class that has the highest predicted probability according to its current state and assumes the class as the true label. Pseudo-labeling is very similar to entropy regularization \cite{entropymin, selftraining}, an older technique in semi-supervised learning. Importantly, in our work, we implement a confidence threshold, which ensures only high-confidence pseudo-labels are used. 

\textbf{Shared Discriminator Architecture}: Many existing methods using GANs for semi-supervised learning utilize a single network with separate classification and discrimination branches \cite{improvedgans, adversarialinference, sslgan}. A traditional classifier attempts to classify data to its respective class, with the output of the classifier being a probability distribution over $K$ such classes. In the popular GAN-classifier methods, the discriminator is given two separate "heads", or final layers, one for classification and one for discrimination. The discriminator output can be represented as the $K+1$-th output, representing the probability of an image being real or fake. We believe this method can be improved on, potentially due to the premise that combining two tasks, discrimination and classification, may reduce performance for both tasks. Since multi-tasking requires the two tasks to be similar \cite{ruder2017overview}, a shared architecture for classification and discrimination may not be optimal because the learned features may be different. We reproduced a shared-architecture (two-headed) GAN to compare and evaluate our results against. 

Our modified loss function for the shared discriminator model can be written as 
\begin{equation}
    \resizebox{0.45\textwidth}{!}{$\lambda(BCE(D_d(D(G(z))), 0) + BCE(D_d(D(x)), 1)) + CE(D_c(D(x)), y)$}
\end{equation} 
where $x$ is real data, $y$ are corresponding labels, $z$ is a randomly generated vector, $\lambda$ is a weight for unsupervised loss, $BCE$ is binary cross-entropy, $CE$ is cross-entropy, $D_d$ is the discriminator head, $D_c$ is the classifier head, $D$ is the discriminator, and $G$ is the generator. In this method, the discriminator head and classifier head each independently update the base network of shared parameters along with their own individual final layers. This means the network attempts to minimize two separate losses with the same parameters, which is our primary concern.

\textbf{Triple-GAN}: Triple-GAN proposes a competing framework for utilizing GANs in classification \cite{triplegan}. The authors of Triple-GAN claim that if a discriminator is given two incompatible tasks, classification and discrimination, the performance of the entire network will decrease. Thus, they utilize a third network, a classifier, for semi-supervised classification. In this framework, the discriminator has the sole task of identifying fake image-label pairs. The classifier and the generator individually create an artificial image-label pair, where the generator creates a fake image with a conditional label, while a classifier generates a pseudo-label for unlabeled data from the training set. The discriminator is then tasked with determining whether a pair has an artificial component, and if so, which component is artificial. 

The primary focuses of our work are inherently different than this approach, as we focus on improving fully-supervised classification, while Triple-GAN primarily focuses on improving GAN generations and secondly attempts to improve semi-supervised learning. In our work, a GAN supplements classifier training, while in Triple-GAN a classifier supplements GAN training. Triple-GAN aims to build a strong discriminator rather than a strong classifier. Most importantly, we focus on increasing dataset size with generated images, whereas Triple-GAN does not train the classifier on generated images at all, which is the primary premise of our work.

\section{Methods}
\label{Methods}

\subsection{EC-GAN}
\label{sec:algorithm}

\begin{figure*}
    \centering
  \includegraphics[width=0.6\linewidth]{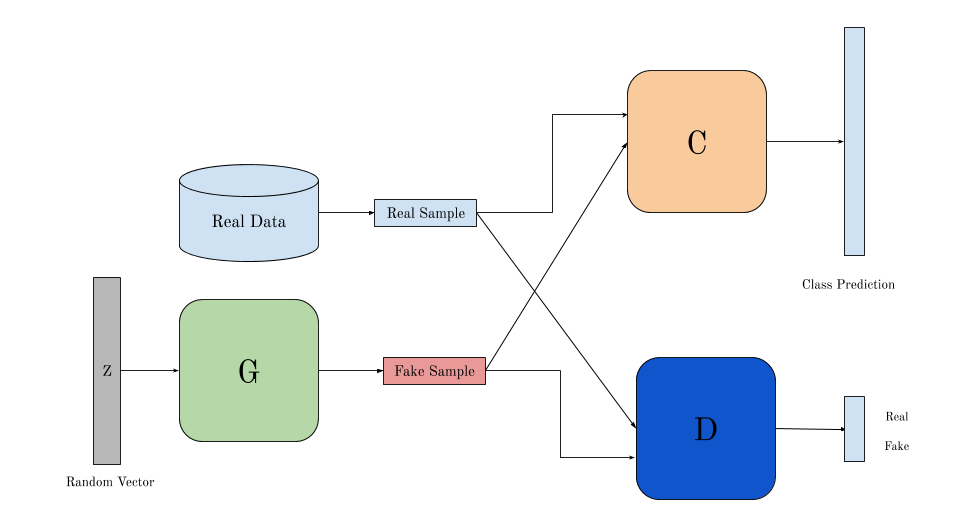}
  \caption{EC-GAN has an external classifier trained on generated images from the generator and real images simultaneously. The discriminator and generator are trained conventionally, and the discriminator's sole output is the predicted probability of the image being real. The datasets are fully-supervised, but the classifier is trained on a semi-supervised algorithm.}
  \label{fig:algorithm}
\end{figure*}

We propose an algorithm to improve classification utilizing GAN-generated images in restricted, fully-supervised regimes. Our approach consists of three separate models: a generator, a discriminator, and a classifier. At every training iteration, the generator is given random vectors and generates corresponding images. The discriminator is then updated to better distinguish between real and generated samples.

Simultaneously, a classifier is trained in standard fashion on available real data and their respective labels (note these datasets are fully labeled, just low-sample datasets). We then use generated images as inputs for supplementing classification during training. This is the semi-supervised portion of our algorithm, as the generated images do not have associated labels. To produce labels for generated images, we use a pseudo-labeling method which assumes a label based on the most likely class according to the current state of the classifier \cite{pseudolabel}. The generated images and labels are only retained if the model predicts the class of the sample with high confidence, or a probability above a certain threshold. Note that new generated images are produced for each minibatch and are immediately fed to the classifier, so no generated images are saved. The loss calculated on generated images and corresponding pseudo-labels is multiplied by a hyperparameter $\lambda$, which controls the relative importance of generated data compared to true samples (described further below). 

Note that this classifier is its own network, as opposed to a shared architecture with the discriminator. This is a key contribution of our paper, as most GAN-based classification methods employ a shared discriminator-classifier architecture. We aim to empirically show that an external classifier performs better than a shared architecture. 

We propose an intuitive loss for our algorithm, which utilizes both supervised and unsupervised methods. As in standard GANs, the discriminator loss is defined by \begin{equation}
    L_D(x, z) = BCE(D(x), 1) + BCE(D(G(z)), 0)
\end{equation}
where BCE is binary cross-entropy, $D$ is the discriminator, $G$ is the generator, $x$ is real, labeled data, and $z$ is a random vector.

The first loss component trains the discriminator on real data, and we compute a loss based on the predictions on the real data and a label of 1, which indicates to the discriminator that the data is real. The second component trains the discriminator on fake images, which inputs generated images and labels of 0, which indicates these images are fake. 

The generator loss can be written as 
\begin{equation}
    L_G(z) = BCE(D(G(z)), 1)
\end{equation}
For the generator, we simply input the predictions of the discriminator and a label of 1 to indicate the predictions are correct. This trains the generator to learn to produce more realistic images.

For our classifier, the loss is defined as 
\begin{equation}
    \resizebox{0.45\textwidth}{!}{$L_C(x, y, z) = CE(C(x), y) + \lambda CE(C(G(z)), {argmax(C(G(z))) > t}) $}
\end{equation}
where $\lambda$ is the unsupervised loss weight (adversarial weight), $CE$ is cross entropy loss, $C$ is the classifier, and $t$ is the pseudo-label threshold. 

The first component of the classifier loss is a standard cross-entropy loss using real data and real labels. The next component is the unsupervised loss. We compute the cross-entropy between generated data and corresponding hypothesized labels. $\lambda$ is an unsupervised weight, and similar approaches have seen success in semi-supervised learning regimes (using real unlabeled data rather than artificially generated data) and can be seen as a regularization parameter \cite{sslweight, entropymin}. Accordingly, we control the extent of regularization using the parameter $\lambda$, which we refer to as the adversarial weight. We incorporate $\lambda$ because generated images are only meant to supplement supervised classification and should be less significant when calculating loss and updating the model parameters. In a scenario where the unlabeled dataset highly outnumbers the labeled dataset, without weighting, the model will inherently under-perform as it is learning less from professional annotations.

For the label in the unsupervised cross-entropy, we use a confidence-based pseudo labeling scheme \cite{pseudolabel}. The pseudo-labeling threshold, $t$, ensures predicted labels are above a specific probability threshold in order to use the image-label pair in loss calculations. GAN images during early training are often low quality, so having a threshold counteracts this. This importantly prevents the model from learning from incorrect and poor labels until the GAN generations are significantly improved. Once the labels are generated, we compute the loss with the artificial label and image and scale it with the adversarial weight.

\subsection{Model Architectures}
The generator, discriminator, and classifier each have an independent architecture in our approach. A random vector of 100 dimensions is inputted into the generator, and after a series of layers and transformations, an image is generated. In certain experiments, we updated the vector to hold class information in the final dimension to condition the generator to generate images of a certain class. We use the Deep Convolutional GAN \cite{dcgan} architecture for our algorithm. The DC Generator architecture uses convolutional-transpose layers, batch normalization layers, and ReLU activation functions. The DC Discriminator uses a similar architecture to the generator, but substitutes convolutional-transpose layers with strided convolutional layers and uses LeakyReLU activations as opposed to ReLU activations. Convolutional layers are employed to follow standard convolutional neural networks (CNNs), making the GAN architecture more suited for image classification. 

\begin{figure}[h]
    \centering
    \begin{minipage}{0.5\textwidth}
        \centering
        \includegraphics[width=0.9\textwidth]{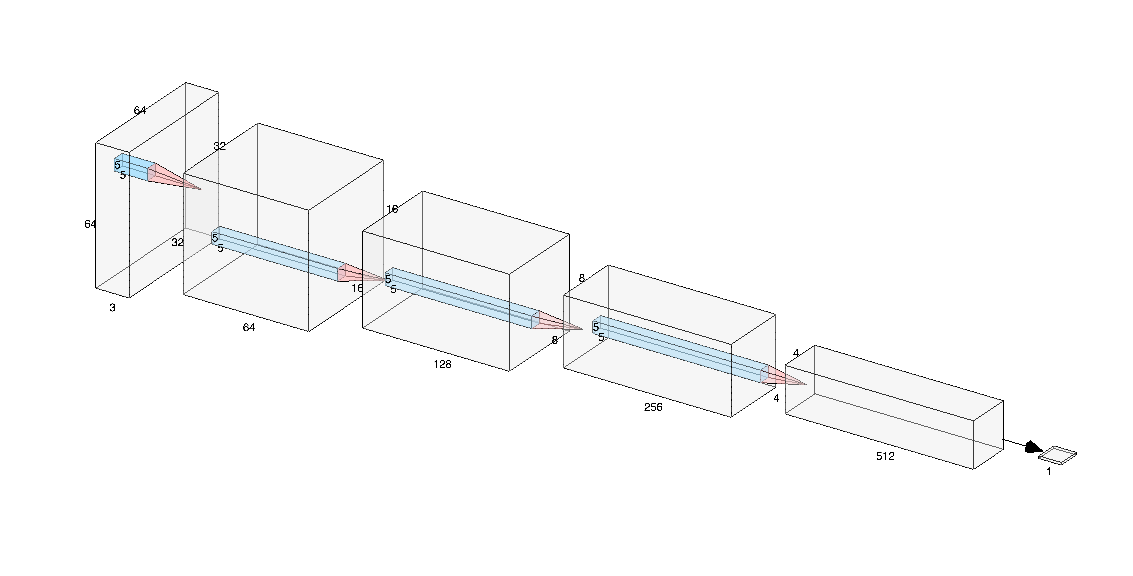} 
    \end{minipage}\hfill
    \begin{minipage}{0.5\textwidth}
        \centering
        \includegraphics[width=0.9\textwidth]{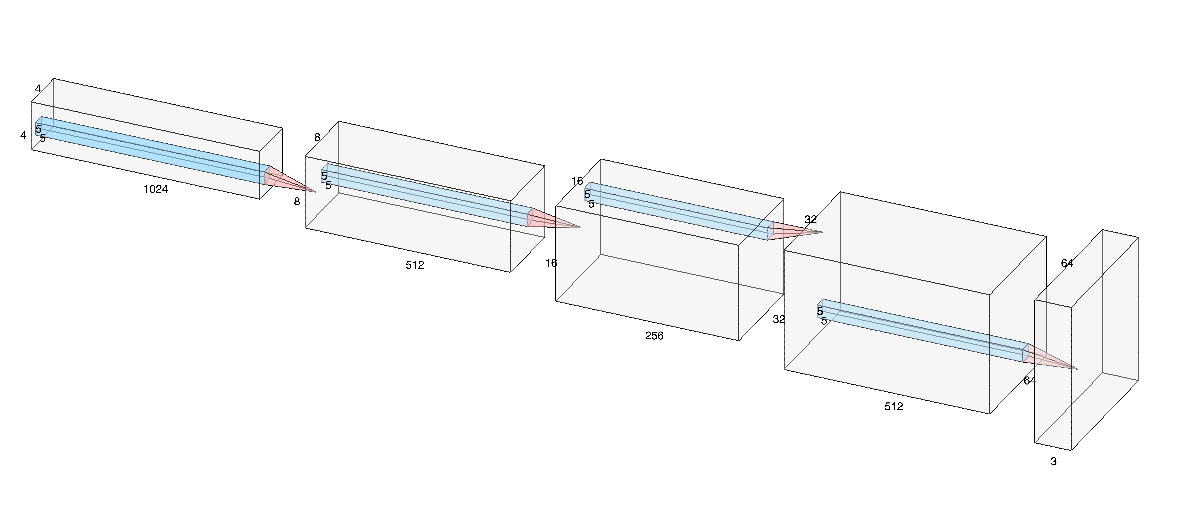} 
    \end{minipage}
    \caption{The architecture of the DCGAN we use is presented. For the discriminator (top), the input image can be of size 64x64 or 32x32. For the generator (bottom), the input is a random vector of 100 dimensions (not shown), and the final output will be a 3-channel image of size 64x64 or 32x32.}
\end{figure}

For our classifier, we used the ResNet-18 neural network, a state-of-the-art image classification model with proven empirical success \cite{resnet}. 

\section{Experimental Evaluation}
\label{Results}

\subsection{Implementation Details}
We implemented our code in PyTorch and used a 12GB NVIDIA Tesla K80 GPU for training. For our algorithm, we used learning rates of 0.0002 (recommended by DCGAN), normalization values of 0.5, and the Adam variant of the SGD optimizer \cite{adam}. For our augmentations, we used a random cropping of 4x4 and random 10-degree rotation. For regularization, we used L2 regularization \cite{weightdecay}, or weight decay, of 0.001. We set $\lambda$ to 0.1, and $t$ to 0.7. Throughout our tests, we manipulated the number of sampled used for training by percent size of the dataset to show the efficacy of our algorithm in small datasets. For all tests, dataset sizes remained relatively small. 

We trained and tested on both an academic dataset and a real-world dataset. We used the Street View House Numbers (SVHN) dataset \cite{svhn} for algorithm development and testing. It is a relatively difficult benchmark dataset due to the images' complex features. The training set is 73,257 images and the validation set is 26,032 images, and the images are 32x32 in size. We varied the number of samples used for training to test the performance at varying dataset sizes. While this dataset is large, it is an important benchmark dataset to evaluate results on, and testing at low amounts of labeled data allows it to fit in our defined goal. 

We also evaluated our method on a Chest X-Ray dataset for pneumonia classification \cite{cxrdataset}. The images were selected from cohorts of pediatric patients from Guangzhou Women and Children’s Medical Center, Guangzhou. All radiographs were screened for quality control and labeled by experts. We downsampled the images before training as GANs empirically struggle with generating large images \cite{biggan}. This dataset contains 5,863 total images, which less than 10\% of the SVHN training set, making the dataset a true scenario of our defined problem statement. For even more severe low-sample scenarios, we varied the amount of training samples on this dataset used for experiments.

\subsection{Shared Architecture Method}
As mentioned, popular algorithms for adversarial training employ a shared, two-headed architecture. EC-GAN is differs because the classifier is external, and this point of contention prompts a comparison in performance of the two methods. The primary hypothesis of our study is that classification and discrimination may be incompatible tasks, which we analyzed through an empirical analysis.

\begin{table*}[]\setlength{\tabcolsep}{4pt}
\centering
\resizebox{\linewidth}{!}{
\begin{tabular}{@{} c c cc c cc c cc@{}}
\toprule
\multirow{2}{*}{Dataset Size (\%)} & \phantom{a} & \multicolumn{2}{c}{EC-GAN (\%)} & \phantom{a} & \multicolumn{2}{c}{Shared DCDiscriminator (\%)} & \phantom{a}& \multicolumn{2}{c}{Shared ResNetDiscriminator (\%)}\\
\cmidrule{3-4}\cmidrule{6-7}\cmidrule{9-10}
&& Classifier & GAN && Classifier & GAN && Classifier & GAN\\
\midrule
10  && 88.63 & 91.15 && 83.54  & 86.17 && 88.63 & 89.32 \\
15  && 90.88 & 92.21 && 85.20 & 88.72 && 90.88 & 91.37\\
20  && 92.61 & 93.40 && 86.77 & 89.39 && 92.61 & 93.24\\
25  && 92.89 & 93.93 && 87.58 & 87.93 && 92.89 & 93.96\\
30  && 93.12 & 94.32 && 87.78 & 90.62 && 93.12 & 93.42\\
\bottomrule
\end{tabular}
}
\caption{EC-GAN is compared to the shared architecture method on SVHN at different dataset sizes. Left value is accuracy of a baseline classifier (same architecture as GAN counterpart), followed by the accuracy of GAN classification algorithm.}
\label{tab:scores}
\end{table*}

Our external classifier method performs on par and occasionally better than a shared architecture in small datasets (Table \ref{tab:scores}). We hypothesize that because each network can learn its own task as opposed to a shared architecture where the network simultaneously fits the representation for two tasks, both networks are able to optimize their own parameters more effectively. Moreover, the shared architecture does not definitionally increase the size of the dataset, since it does train classification on GAN images. Through our empirical analysis, we contend that separating classification and discrimination and supplementing classification with generated images are key factors for strong classification performance. 

\subsection{Conventional Approaches}
In restricted data regimes, data augmentation and regularization are both conventional methods leveraged to improve accuracy and generalization. We evaluated against data augmentation and regularization to find practical value in our algorithm. We tested our algorithm with no strategies against the baseline with each strategy.

\begin{figure}[h]
    \centerline{\includegraphics[scale=0.4]{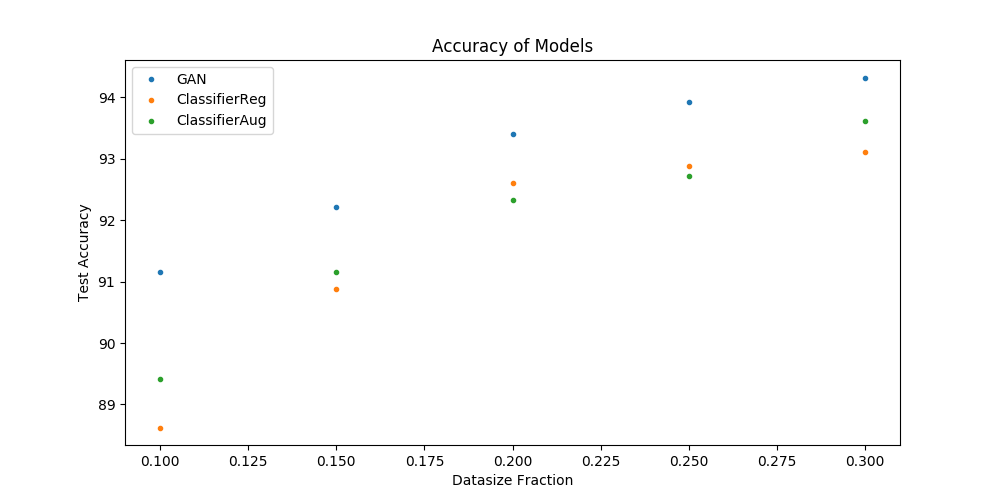}}
    \caption{EC-GAN with no additional strategies compared to a standard classifier (same architecture) with either regularization or data augmentation.}
    \label{fig:conventionalapproaches}
\end{figure}

EC-GAN performs better than standard data augmentation and regularization procedures (Figure \ref{fig:conventionalapproaches}). Regularization has long been a standard procedure to increase generalization of a model, but by increasing a dataset with new and unique images, generalization of a model may increase to a higher degree with our method. Data augmentation only uses existing images and makes different transformations, so in essence, they are still already existing data. However, a GAN generates new images, meaning with our algorithm, a dataset is enlarged to a higher degree. EC-GAN outperforms conventional practices, supporting the viability and effectiveness of our novel approach.

\subsection{X-Ray Dataset}
This dataset has a strong class imbalance, so GAN generated images will reflect this imbalance. We utilized a conditional GAN \cite{conditionalgan} in order to generate evenly balanced fake images. We modified our random vector to encode balanced class information in the last dimension. For our discriminator, we concatenate an extra channel in a hidden layer with class information, which allows the discriminator and generator to learn class-specific characteristics. This method proves to have better results.

\begin{table}[h]
\setlength{\tabcolsep}{4pt}
\centering
\resizebox{0.6\linewidth}{!}{
\begin{tabular}{@{} c c ccc@{}}
\toprule
\multirow{2}{*}{Dataset Size (\%)} & \phantom{a} & \multicolumn{3}{c}{EC-GAN (\%)} \\
\cmidrule{3-5}
&&  Classifier && GAN \\
\midrule
25 &&  94.37 && 96.48 \\
50  && 95.24 && 97.83 \\
75  && 95.64 && 97.40 \\
100  && 96.42 && 97.99 \\
\bottomrule
\end{tabular}
    }
\caption{The conditional version of EC-GAN is tested on the X-ray dataset. The left value is the accuracy of the baseline classifier and the right value is the accuracy of EC-GAN.}
\label{tab:scores2}
\end{table}

The results of this experiment (Table \ref{tab:scores2}) reflect the success of EC-GAN. Our approach strongly improves classification, especially in realistic and small datasets. To test an extreme case, we trained with just 100 images per class and achieved an accuracy of 90.9\%, which is impressive for such a low-sample training task of 200 images.

\section{Ablative Results}
The following section will investigate ablative results and miscellaneous experimentation.

\subsection{Standard GAN}
\begin{figure}
    \centering
    \begin{tabular}{ll}
    \includegraphics[scale=0.35]{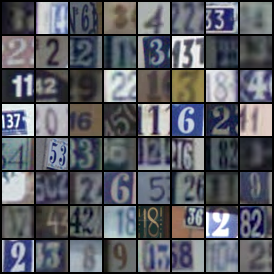}
    &
    \includegraphics[scale=0.35]{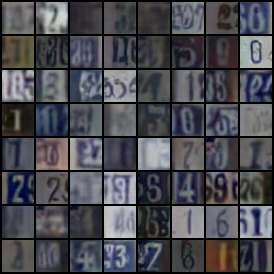}
    \end{tabular}
    \caption{Left: Real SVHN images.
    Right: Fake SVHN Images. The fake SVHN images are generated with EC-GAN. Most images resemble real images, as there are distinguishable digits and key features are being generated.
    }
    \label{Fig:SVHN}
\end{figure}

\begin{figure}
    \centering
    \begin{tabular}{ll}
    \includegraphics[scale=0.35]{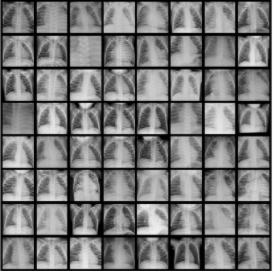}
    &
    \includegraphics[scale=0.1792]{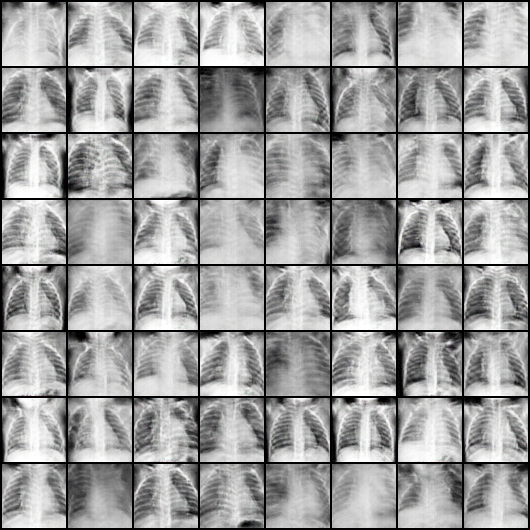}
    \end{tabular}
    \caption{Left: Real x-ray images.
    Right: x-ray Images. The fake x-ray images are generated with EC-GAN. Most images are clear and there are distinct features such as rib cages.
    }
    \label{Fig:CXR}
\end{figure}

To begin the experimentation process, we implemented a DCGAN without any classifier. We evaluated the quality of the fake images generated by the GAN to ensure accurate and proper images were used for training (Figures \ref{Fig:SVHN} and \ref{Fig:CXR}). Tests were conducted on both the SVHN and the x-ray dataset. 

\subsection{Unsupervised Loss Weight (Adversarial Weight)}

After initial testing of our algorithm, we found our method was performing worse than the baseline classifier at all dataset sizes. We found, especially early during training, the accuracy of the model was reduced by GAN generations because the relative importance of the unsupervised loss ($\lambda$) was not defined. Since most semi-supervised algorithms utilize a weighted unsupervised loss, we investigated the strategy (explained in loss functions section). 

\begin{figure}[h]
    \centerline{\includegraphics[scale=0.6, ]{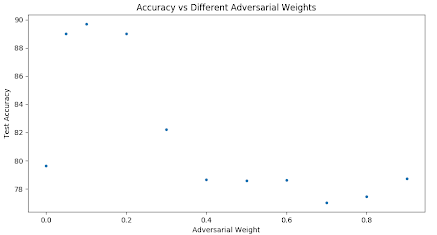}}
    \caption{The plot illustrates a significant increase in accuracy after introducing weighted unsupervised loss ($\lambda$). Plot shows optimal value of $\lambda$ is 0.1.}
    \label{fig:advPlot}
\end{figure}

The accuracy increases considerably with an adversarial weight, as EC-GAN with $\lambda$ outperforms the baseline classifier (Figure \ref{fig:advPlot}). With just an adversarial weight of 0.1, the accuracy of our model increases by over 10 percentage points. Using an adversarial weight allows for effective usage of GAN generated images, as generated or unlabeled images generally will not be as beneficial for a classifier as real and labeled images because of their faulty and somewhat unreliable nature. 

\subsection{Additional Training Procedures}
We tested multiple different strategies in order to improve performance of the algorithm. We tuned specific hyperparameters for each strategy and compared the performance of the strategies against one another. The tested strategies include L2 regularization and data augmentation.

\begin{figure}[h]
    \centerline{\includegraphics[scale=0.3]{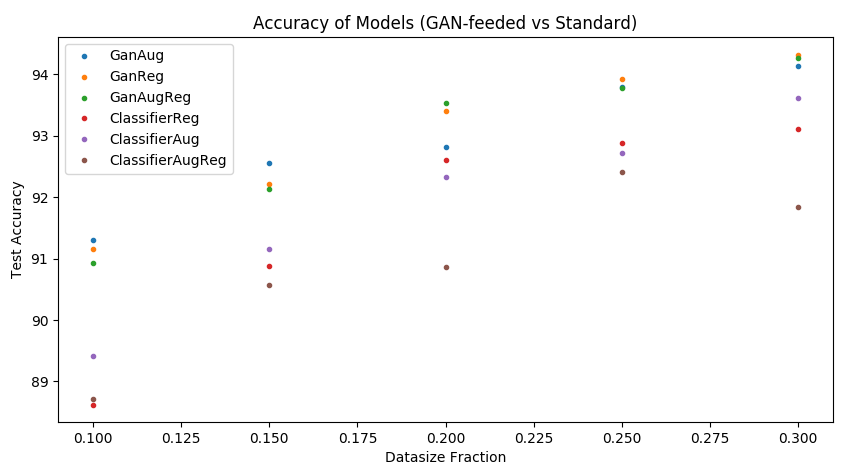}}
    \caption{EC-GAN is trained with data augmentation, regularization, and both. The same training procedures are applied to the baseline classifier, and results of both are compared and displayed.}
    \label{fig:strategies}
\end{figure}

Even with the additional training procedures, our proposed method improves the performance of all models (Figure \ref{fig:strategies}). We find that EC-GAN performs best with regularization. Augmentation has similar performance improvements, but the addition of regularization provides more consistency across dataset sizes, and therefore includes an L2 regularization procedure in our algorithm.

\section{Conclusion}
We have presented EC-GAN, a novel generative model that attaches an external classifier to a GAN to improve classification performance in restricted, fully-supervised datasets. Our proposed method allows classifiers to leverage GAN image generations to improve classification, while simultaneously separating the tasks of discrimination and classification. Our results showed that EC-GAN is effective and can be used to improve image classification performance in small, real-world datasets. Our future work will focus on evaluating performance in different domains as well as developing new semi-supervised methods, such as psuedo-labeling with a conditional-GAN. 

\small
\bibliography{main}

\end{document}